\documentclass{article}

\usepackage{arxiv}
\usepackage{tabularx}
\usepackage{array}
\newcolumntype{L}[1]{>{\raggedright\arraybackslash}p{#1}}
\usepackage{amssymb}

\usepackage{tikz}
\usepackage[utf8]{inputenc}
\usepackage[T1]{fontenc}
\usepackage{hyperref}
\usepackage{url}
\usepackage{booktabs}
\usepackage{amsfonts}
\usepackage{amsmath}
\usepackage{nicefrac}
\usepackage{microtype}
\usepackage{graphicx}
\usepackage{natbib}
\usepackage{doi}
\usepackage{algorithm, algorithmic}
\usepackage{longtable}

\newcommand*\emptycirc[1][1ex]{\tikz\draw (0,0) circle (#1);} 
\newcommand*\halfcirc[1][1ex]{
  \begin{tikzpicture}
  \draw[fill] (0,0)-- (90:#1) arc (90:270:#1) -- cycle ;
  \draw (0,0) circle (#1);
  \end{tikzpicture}}
\newcommand*\fullcirc[1][1ex]{\tikz\fill (0,0) circle (#1);}

\usepackage{url}

\usetikzlibrary{positioning, chains}

\usepackage{soul}

\newcommand{\colorhighlight}[3]{
  \begingroup
    \colorlet{hlcolor}{#1!#2!white}
    \sethlcolor{hlcolor}
    \hl{#3}
  \endgroup
}

\title{Stylometric Watermarks for Large Language Models}

\author{ \href{https://orcid.org/0009-0008-9569-1313}{\includegraphics[scale=0.06]{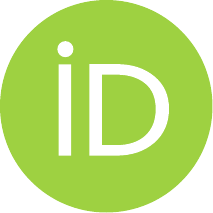}\hspace{1mm}Georg Niess} \\
	Institute of Interactive Systems and Data Science\\
	Graz University of Technology\\
	Graz, Austria \\
	\texttt{georg.niess@tugraz.at} \\
	\And
	\href{https://orcid.org/0000-0003-0202-6100}{\includegraphics[scale=0.06]{orcid.pdf}\hspace{1mm}Roman Kern} \\
	Know-Center GmbH \&\\
    Institute of Interactive Systems and Data Science\\
	Graz University of Technology\\
	Graz, Austria \\
	\texttt{rkern@know-center.at} \\
}

\renewcommand{\shorttitle}{Stylometric Watermarks for LLMs}

\hypersetup{
pdftitle={Stylometric Watermarks for Large Language Models},
pdfsubject={cs.CL},
pdfauthor={Georg Niess, Roman Kern},
pdfkeywords={Watermark, Stylometry, Accountability, LLM},
}

\begin{document}
\maketitle

\begin{abstract}
    The rapid advancement of large language models (LLMs) has made it increasingly difficult to distinguish between text written by humans and machines.
    Addressing this, we propose a novel method for generating watermarks that strategically alters token probabilities during generation.
    Unlike previous works, this method uniquely employs linguistic features such as stylometry. 
    Concretely, we introduce acrostica and sensorimotor norms to LLMs. 
    Further, these features are parameterized by a key, which is updated every sentence.
    To compute this key, we use semantic zero shot classification, which enhances resilience.
    In our evaluation, we find that for three or more sentences, our method achieves a false positive  and false negative rate of $0.02$.
    
    For the case of a cyclic translation attack, we observe similar results for seven or more sentences.
    This research is of particular of interest for proprietary LLMs to facilitate accountability and prevent societal harm.

\end{abstract}


\section{Introduction}
Large language models (LLMs) made major improvements with the inception of encoder and decoder models using attention in the form of transformers \citep{vaswani_attention_2017}. Further refinements were made to LLMs using model pre-training \citep{devlin_bert_2019} and very large data sets \citep{radford_language_2019}. Through these developments, it has continuously become harder for humans to differentiate texts written by LLMs from those written by humans, with machine-generated texts sometimes even fooling humans more often than human-written texts \citep{zellers_defending_2019}. While machine detection techniques exist, they are slowly being outgrown by the progress of LLMs, for example, several detection methods for GPT-2 are already shown to struggle with GPT-3 \citep{fagni_tweepfake_2021}.
This trend already causes worries in many aspects that will only increase with the trend of improvement and consistent scaling of language models \citep{kaplan_scaling_2020}.
Several potential cases of misuse already exist for these advanced models \citep{ray_chatgpt_2023}. 

\begin{figure*}[ht]
    \vskip 0.2in
    \begin{center}
    \centerline{\includegraphics[page=1, width=0.9\textwidth]{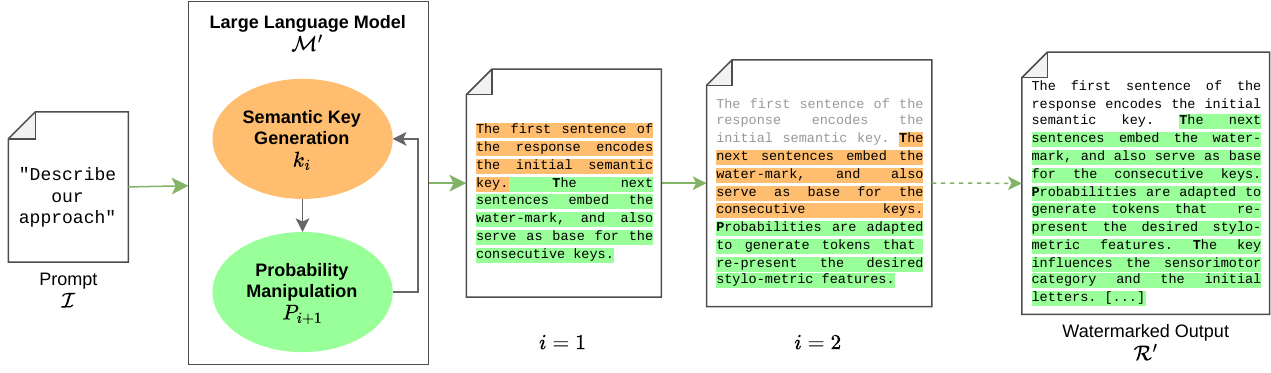}}
    \caption{Overview of our method, where during the text generation a semantic key is derived from each sentence (orange highlight). The key controls the update of the probabilities of the generated tokens in the following sentence to reflect stylometric features, which represent our watermark (green highlight). The detection of the watermark will work for any sequence of sentences from a longer document (i.e., not limited to an entire response). 
    }
    \label{fig:overview}
    \end{center}
    \vskip -0.2in
\end{figure*}

To address these issues, we propose an approach for injecting a watermark into the output of a large language model, which then allows testing for the presence of the watermark with a high degree of certainty.
The main concept of our approach is to update the probabilities of the generated tokens directly during the generation process, following works like~\citep{kirchenbauer_watermark_2023}. 
In contrast to existing works, we focus on stylometric features.
Another central aspect of our method is to additionally control these features via a key. 
The key changes dynamically and is derived directly from the output of the language model, based on the semantics of the sentence.
Concretely, the first sentence of the response as generated by the LLM encodes the initial key used to control the stylometric watermark of the next sentence.
This sentence again is the input to derive the next, consecutive key.
Figure~\ref{fig:overview} presents a conceptual overview of our approach together with alternative approaches.
Our method is flexible to allow for a diverse set of features, while in this work we limit the features to just sensorimotor norms and acrostica.

Stylometry is the study of assessing a person's writing style, with many stylometric features being proposed in literature \cite{neal_surveying_2017}. 

These features are like a writer's fingerprint, including syntax, the vocabulary used, structure and size of sentences and other idiosyncrasies of the author. 
All these features can be detected statistically and then be used in tasks like authorship attribution~\citep{stamatatos2009survey}.
However, classic authorship attribution has shown to struggle with smaller LLMs, such as GPT-2~\cite{uchendu_authorship_2020}. 
Our proposal of using a key to deliberately alter the style per sentence can increase the ability to detect stylometric features to make them statistically significant.

Sensorimotor norms are categories based on human cognition. 
Perceptual modalities like "hearing" or action effectors like "hand" are well researched in psychology and cognitive semantics~\cite{lynott_lancaster_2020}, but there is yet little research in computer science.
In our approach, the key selects the sensorimotor category and thus influences the generated words, e.g., for the olfactory category "smells funny" would be preferred over "looks funny".

An acrostic is a text where the first letter in each sentence can be combined to spell out a secret message or word. 
Historically, acrostica have been used by authors by encoding their authorship~\cite{johnson2006authorial} with variations of their names being the hidden acrostic. An famous example for an acrostic is reported in Appendix~\ref{tab:acrostic-schwarz}.
In our approach, the key controls the letters to be used as the first letter of the first word of a generated sentence.

The main contributions presented in this paper are:

\begin{itemize}
    \item We propose a novel watermarking approach for large language models based on changing the probabilities on a sentence-based level to embed stylometric features.
    \item The method does not require a dedicated fine-tuning training dataset, and does not require an LLM for checking for the watermark.
    \item We are the first to introduce sensorimotor words to encode a linguistic style into generated text via LLMs.
    \item To derive resilient keys, we introduce a fuzzy hashing based on semantic labelling via zero-shot classification.
    \item To detect watermarks we make use of statistical hypothesis tests and demonstrate the high effectiveness of the approach and its resilience against attacks, including cyclic translation.
\end{itemize}

\section{Background \& Related Work}
\begin{table*}[htbp] 
\small 
\caption{Overview of approaches suitable for watermarking, together with a list of stylometric features. For each combination, the suitability is self-assessed. A black circle shows the best compatibility, followed by the semi-filled circle. 
Updating the probabilities provides great flexibility and allows for features such as acrostics. The right side presents the architecture of the approaches.}
\begin{minipage}[b]{0.45\linewidth}
    \begin{tabular}{lcccc}
    \toprule
    \textbf{Feature Type} & \textbf{Probability} & \textbf{Fine-Tuning}  & \textbf{Prompt} & \textbf{Post-Processing}  \\
    \midrule
    N-Grams                 & \fullcirc & \fullcirc & \emptycirc & \emptycirc \\
    Character Frequency     & \fullcirc & \halfcirc & \halfcirc & \halfcirc \\
    Vocabulary Richness     & \fullcirc & \fullcirc & \halfcirc & \fullcirc \\
    Word Distributions      & \halfcirc & \fullcirc & \halfcirc & \halfcirc \\
    Word Length             & \fullcirc & \fullcirc & \halfcirc & \halfcirc \\
    Sentence Length         & \halfcirc & \fullcirc & \emptycirc & \emptycirc \\
    Parts of Speech         & \emptycirc& \fullcirc  & \emptycirc & \emptycirc  \\
    Punctuation Frequency   & \halfcirc & \halfcirc & \emptycirc & \halfcirc  \\
    Sentence Complexity     & \halfcirc & \fullcirc  & \halfcirc & \halfcirc   \\
    Synonyms                & \emptycirc& \halfcirc  & \emptycirc & \fullcirc  \\
    \midrule
    Sensorimotoric Words    & \fullcirc & \fullcirc & \fullcirc & \emptycirc  \\
    Acrostics               & \fullcirc & \emptycirc& \halfcirc & \fullcirc  \\
    \bottomrule
    \end{tabular}\\
\end{minipage}
\hspace{1cm}
\begin{minipage}[b]{0.45\textwidth}
    \hspace{3.5cm}
    \vspace{-.4cm}
    \includegraphics[width=4.5cm]{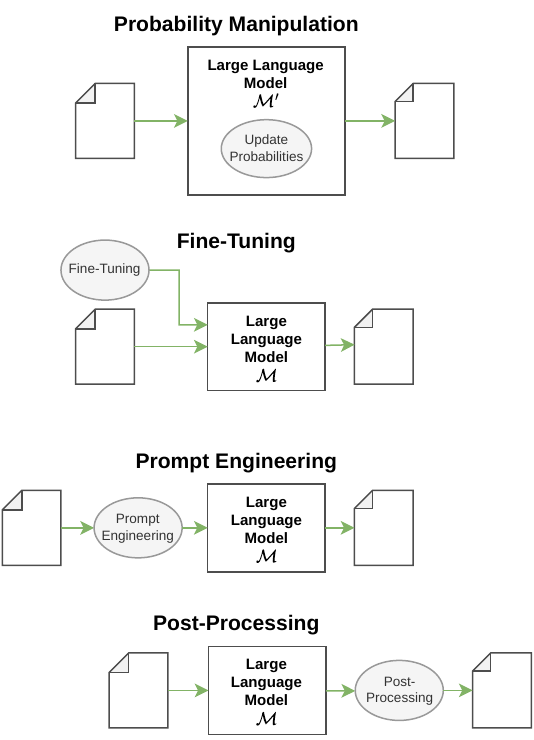}\\
\end{minipage}
\label{table:stylometricFeaturesCompatibility}
\end{table*}

Several notable attempts have already been made to detect machine-generated text, albeit without incorporating language and stylometric features like us. These efforts can generally be categorized into two primary groups, depending on their functional approach.

\paragraph{Post-hoc detection.}In this case, it is assumed that the LLM has already completed text generation some time in the past. The methods in this group attempt to detect the generated text without employing any watermark or other modifications to either the language model or the final output. Notable examples of this group are the classifier by OpenAI~\cite{kirchner_new_2023}, GPTZero~\cite{tian_gptzero_2023} or DetectGPT~\citep{mitchell_detectgpt_2023} who uses the probability curvature of text sampled from a LLM. DetectGPT utilizes the property that text tends to occupy negative curvature regions of the log probability of the model. This implies that any minor modifications to a sentence will result in a reduction in its log likelihood. This is because an LLM continuously strives to attain the optimal probability for each sentence. More post-hoc methods can be found in the survey by~\citet{jawahar_automatic_2020}.

\paragraph{Watermarking.}
Watermarking is the technique of hiding information in data that is difficult to remove by others but can be detected by an algorithm to read the hidden information. 
Some successful watermark implementations already exist, such as \cite{kirchenbauer_watermark_2023}, \cite{christ_undetectable_2023}, and there is ongoing research by \cite{aaronson_my_2022, aaronson_watermarking_2023}. 
Specifically, \cite{kirchenbauer_watermark_2023} separates tokens into a green and red list, where tokens in the green list get a weight boost to increase their representation in the output. This leads to generated text mostly consisting of words from the green list, while human written text naturally also uses words from the red list. 
\citet{xiang2024reversible} address the issue of replacing sensitive words and make use of Sentence-BERT~\citep{reimers2019sentence}.

Post-hoc detection methods have the benefit that they can be used on any suspected texts without any prerequisites, but they are susceptible to break by user attacks \cite{kirchenbauer_reliability_2023}, and it becomes more difficult to use them with complex language models \cite{chakraborty_possibilities_2023}. Watermarks, on the other hand, can be more resilient to attacks by the user, but to enforce their use requires language  models to be mostly centralized by large closed source vendors. Since many contemporary LLMs are proprietary operated by large companies, it opens the possibility to implement them comprehensively, either voluntarily or by regulation.
This may also be an essential part to achieving a desired level of accountability.

\paragraph{Stylometric Features.}
There is a broad range of stylometric features previously proposed in literature \citep{lagutina_survey_2019,stamatatos_survey_2009}. 
They can broadly be categorized into five levels: lexical, syntactic, semantic, structural and domain-specific features. 
In Table~\ref{table:stylometricFeaturesCompatibility} we collected the most popular stylometric features in literature as well as additional watermark features that are promising.

\paragraph{Sensorimotor Norms.}
In \citep{winter2019sensory} they define the term sensory linguistics as the study of how senses and languages are related.
The Lancaster Sensorimotor Norms~\citep{lynott_lancaster_2020} describes a set of 40,000 sensorimotor words together with a crowed-based assessment into 11 dimensions, which also form the base of our sensorimotor features.
Words can be represented by six perceptual groups (touch, hearing, smell, taste, vision, interoception) and five actions (mouth/throat, hand/arm, foot/leg, head excluding mouth/throat, and torso).
\citet{khalid2022smells} make use of the Lancaster Norms and find that sensorial language might be intentionally used and cannot be considered a random phenomenon.
Perceptional features have also been proposed in literature for authorship attribution, specifically in a cross-language setting~\citep{bogdanova2014cross}.

\paragraph{Acrostica.}
\citet{stein2014generating} considered the task of generating a paraphrased version of an existing text, where the generated text should contain an acrostic.
Thereby, they formulated this as a search problem.
\citet{shen2019controlling} utilised a sequence-to-sequence network to generate text incorporating acrostica for English and Chinese.
More recently, steganography has been considered to embed secret messages in text using BERT~\citep{yi2022alisa}.
\subsection{Watermarking Approaches}

Watermarks, including but not limited to stylometric ones, can be utilized in several ways by LLMs. In the following, we present four different approaches. 
They have different strengths and weaknesses we provide an overview and estimate how effective each approach is expected to be for different stylometric features  in Table~\ref{table:stylometricFeaturesCompatibility}.

\paragraph{Probability Manipulation.}
 In our approach, we have opted for directly manipulating the probabilities of the generated tokens of the LLM. 
 This gives a lot of control to how single features behave, though the difficulty lies in finding the correct probabilities that produce exactly that desired feature. 
 An example of a watermark of this type is the work by \citet{kirchenbauer_watermark_2023}. 

\paragraph{Fine-tuning.}
To change the output of an LLM, a common solution is to conduct additional training. This, of course, also works with watermarks, through it is less controlled than other methods. For example, the written material of an author with a known writing style could be used to fine-tune an LLM to produce text similar to that author~\cite{li_teach_2023}. 
The problem is that it is difficult to control what is learned and what not. 
The model might learn not only the wanted watermark features, but also overfit to other features that can lead to a loss of generalization, domain mismatch and in general reduced robustness for outputting text.

\paragraph{Prompt Engineering.}
Another option is to use the innate capability of LLMs and to engineer a system prompt that tells the model what features are desired~\cite{zhou_large_2023}. For example, it is possible to tell contemporary LLMs to use only certain letters in its sentence~\cite{openai_chat_2024b}. The challenges here are creating the right prompt, defending against users writing their own prompt to trick this, and what prompts the language model can even comprehend and adhere to. In general, a more powerful model is more suitable for this method.

\paragraph{Post-Processing.}
The last possibility is to post process the text after generation is finished. This is how early attempts were done by \citet{topkara_natural_2005} and \citet{atallah_natural_2001} when LLMs were not yet developed. This was popular for embedding watermarks at the time, for example for copyright or document integrity. Although the generative power of LLMs has made this less attractive, it can still be of use for some simple features like synonym replacement that do not require the capability to generate new sentences.

\section{Notation and Prerequisites}
This section defines prerequisite notations for later use in this work.
A large language model \(\mathcal{M}\) is for our purpose a generative transformer that takes some user input prompt \(\mathcal{I}\) and outputs a response \(\mathcal{R}\). 
The response consists of $n$ sentences, and $s_i$ indicating the i-th sentence in the response.
A token is what \(\mathcal{M}\) can generate in one iterative step. We define our vocabulary of possible tokens as \(\mathcal{V}\). \(\mathcal{M}\) does so by combining \(\mathcal{I}\) with the already generated output tokens to generate a sentence-specific probability distribution \(\mathcal{P}_i\) over all tokens in \(\mathcal{V}\). Then an algorithm, for example a greedy or beam search, chooses a token with sufficiently high probability to use as the next output token. This token is fed into  \(\mathcal{M}\) again combined with \(\mathcal{I}\) and the other previous output tokens to continue sampling \(\mathcal{P}_i\) until a stop token is chosen.

The next step is to define a watermarked language model \(\mathcal{M'}\). First, in the embedding step, a watermark \(\mathcal{W}\) is embedded into \(\mathcal{R}\) during the generation process. This is done by manipulating the probability distribution \(\mathcal{P}_i\) in some way, depending on \(\mathcal{W}\). This leaves us with our watermarked output response \(\mathcal{R'}\). This means that instead of choosing the token that would have normally been selected, another token is chosen that allows us to change the generated text in some way to build our watermark.  Last, in the detection step, we need to detect \(\mathcal{W}\) in \(\mathcal{R'}\) by extracting the hidden information and calculating how probable it is to be our \(\mathcal{W}\).

\section{Method} 

Our approach consists of two main parts: 1) the watermark generation process based on manipulation of probabilities controlled via dynamic keys, and 2) a test procedure for detecting an existing watermark based on statistical tests.

\subsection{Watermark Generation} \label{construction}

The generation of the watermark can be further split into two separate processes.
First, a key is generated that controls the second process, the manipulation of the probabilities of the tokens.
In Algorithm~\ref{alg:ellm} the key steps of the entire generation are presented.

\begin{algorithm}
\caption{Watermark Generation Algorithm}\label{alg:ellm}
\begin{algorithmic}[1]
\REQUIRE Language model $\mathcal{M'}$, input prompt $\mathcal{I}$, stop token
\ENSURE Watermarked response $\mathcal{R'}$
\STATE Initialize language model $\mathcal{M'}$
\STATE $\mathcal{I}$ $\leftarrow$ prompt
\STATE $\mathcal{V} \leftarrow \mathcal{M'}$\texttt{.get\_vocab()}  
\STATE \textcolor{blue}{\# Generate initial token}
\STATE $t$ $\leftarrow$  \texttt{$\mathcal{M'}$.generate\_next\_token($\mathcal{I}$)}
\STATE Initialize response $\mathcal{R'} \leftarrow ""$
\WHILE{$t \neq$ \texttt{stop\_token}}
\STATE \textcolor{blue}{\# Iterate vocabulary, mark tokens that match key}
\STATE $m \leftarrow$ \texttt{build\_mask($\mathcal{V},\mathcal{K}$)}
\STATE \textcolor{blue}{\# Manipulate weights of marked tokens}
\STATE $\mathcal{P}_i \leftarrow$ \texttt{$\mathcal{P}_i$.boost\_tokens($m$)}
\STATE \textcolor{blue}{\# Generate next token using the modified weights}
\STATE $t$ $\leftarrow$  \texttt{$\mathcal{M'}$.generate\_next\_token($\mathcal{R'}$,$\mathcal{P}_i$)}
\IF{$t = $ \texttt{start\_of\_new\_sentence}}
        \STATE \textcolor{blue}{\# Get the last sentence from the cumulative reply}
        \STATE $s \leftarrow$ \texttt{get\_last\_sentence($\mathcal{R'}$)}
        \STATE \textcolor{blue}{\# Update key with semantic zero shot classification}
        \STATE $\mathcal{K}$ $\leftarrow$ \texttt{get\_semantic\_label($s$)}
    \ENDIF
\STATE $\mathcal{R'} \leftarrow \mathcal{R'} + t$
\ENDWHILE
\STATE Watermarked reply is $\mathcal{R'}$
\end{algorithmic}
\end{algorithm}

\paragraph{Semantic Key Generation.}

A central idea of our approach is to vary the watermark throughout the text.
For this to work, we propose to update the key with each generated sentence.
This key is used to parameterize the stylometric features for each sentence. 
More formally, each sentence $s_i$ will yield a key $k_i$, which controls the token probabilities $P_{i+1}$ of the following sentence $s_{i+1}$.
We propose to use full preceding sentences for deriving the keys, but other approaches like a sliding window are also possible, but kept for future research. 

We note that the naive approach of using a simple hash as key being computed from previous tokens is susceptible to attacks.
Any manipulation of the previous sentence may change the key, and thus the ability to detect the watermark will be impeded.
To increase the resilience of the key generation, we propose a fuzzy hashing approach based on the semantics of the sentence.
Our approach is not limited to a specific type of key generation.
As such, any provider of LLMs may choose to use their unique version of key generation.

Our implementation uses Bart~\cite{lewis_bart_2019} trained on the MultiNLI dataset \cite{williams_broad-coverage_2018} as a zero shot classifier of sentences \cite{yin_benchmarking_2019}. 
We created labels with the help of ChatGPT~\cite{openai_chat_2024a} representing broad coverage of semantics to be able to have a generalized classification of most sentences. 
Then, the key is constructed depending on the chosen label for each stylometric feature. 
For example, since there are 11 sensorimotor groups, 11 labels are created and numbered.
Bart will then be used to assign the key sentence to one of these 11 labels.
For a complete list of the labels and their corresponding features, see Appendix \ref{table:labels}.

\paragraph{Probability Manipulation.}
With the current key, a new sentence can now be constructed. First, the key is used to decide how the watermark features are to be parameterized. Then, all tokens in the embedding space are loaded. The model can now look up the next token it would generate by calculating the weight for each token in the embedding space. Normally, some sampling algorithm would now choose which token would be selected, based on the probability distribution \(\mathcal{P}_i\). 
However, in our case, the tokens are now converted from the embedding space to letters and checked against our selected stylometric features. If a token depicts our desired feature, either in part or in full, it is marked in a mask spanning all tokens in the embedding space. 

This mask is then used to slightly shift the probabilities of the tokens, based on a weight factor.
This factor should be relatively small as to not influence the output of the model too much. 
The sampler should still be allowed to choose highly probable tokens, not impeding the semantics or fluency of the generated output. 
An example would be that after the word “Barack”, it is most likely to have the word “Obama”. 
It is preferable that the sampler is represented with multiple options with close probabilities.
Thus the probability manipulations should be taken with great care.
Each feature should also have different weight factors depending on its sensibility.
While a wide variety of features can be embedded by this approach, we describe two feature types in the following. 

\paragraph{Acrostic.} An acrostic is a text where the first letter in each sentence can be combined to spell out a secret message or word. In our case, we do not require all first letters to follow this characteristic.
Depending on text length, even a few matching letters can lead to very high confidence, which allows the model to be more free in token selection and also improves resilience.

We chose a relatively large weight boost for acrostica since the beginning of the sentence is more flexible and the rest of the sentence can easily adapt to this change. 
Static weight factors for those two features worked well in our experiments, but weights could also be chosen dynamically depending on how long it has been since a desired feature was chosen or by how the distribution of the current weights looks like.

\paragraph{Sensorimotor Norms.}
 We used a large set of semantic norms \cite{lynott_lancaster_2020} to be able to cross-reference our vocabulary. A much smaller weight was used for this feature compared to the acrostica since a larger weight could lead to a lot of bias in the generative capabilities of the models, and since we do not only consider the first word of the sentence, but all of them now, it is not so essential to have a strict selection according to what the feature dictates.
For sensorimotor words, we chose a small weight so that the model would not be overly biased. 

\begin{table*}[ht]
\scriptsize
\centering
\begin{tabular}{p{2.1cm}|p{3.2cm}|p{3.2cm}|p{3.2cm}|p{3.2cm}}
\toprule
\textbf{Prompt}  & \textbf{Base with no watermark} &  \textbf{Watermark with sensorimotoric only} &  \textbf{Watermark with acrostic only} & \textbf{Watermark with both}  \\
\midrule
    What is motorsports, is it a real sport?  
    & 
    Motorsports is a sport that uses vehicles for competition.
    \colorhighlight{red}{0}{Many}\colorhighlight{red}{63.52941176}{people}\colorhighlight{red}{0}{consider}\colorhighlight{red}{0}{motorsports}\colorhighlight{red}{0}{to}\colorhighlight{red}{0}{be}\colorhighlight{red}{0}{a}\colorhighlight{red}{0}{form}\colorhighlight{red}{0}{of}\colorhighlight{red}{0}{racing}\colorhighlight{red}{0}{but}\colorhighlight{red}{0}{they}\colorhighlight{red}{0}{can}\colorhighlight{red}{0}{also}\colorhighlight{red}{0}{be}\colorhighlight{red}{0}{classified}\colorhighlight{red}{0}{as}\colorhighlight{red}{0}{an}\colorhighlight{red}{0}{extreme}\colorhighlight{red}{0}{sport}\colorhighlight{red}{0}{due}\colorhighlight{red}{0}{to}\colorhighlight{red}{0}{the}\colorhighlight{red}{0}{level}\colorhighlight{red}{0}{of}\colorhighlight{red}{0}{danger}\colorhighlight{red}{0}{and}\colorhighlight{red}{0}{physical}\colorhighlight{red}{0}{exertion}\colorhighlight{red}{0}{involved.}
    \colorhighlight{red}{0}{The}\colorhighlight{red}{0}{term}\colorhighlight{red}{0}{motorsport}\colorhighlight{red}{0}{is}\colorhighlight{red}{0}{used}\colorhighlight{red}{0}{to}\colorhighlight{red}{0}{refer}\colorhighlight{red}{0}{to}\colorhighlight{red}{0}{many}\colorhighlight{red}{0}{different}\colorhighlight{red}{0}{types}\colorhighlight{red}{0}{of}\colorhighlight{red}{0}{competitions}\colorhighlight{red}{0}{involving}\colorhighlight{red}{0}{motorized}\colorhighlight{red}{0}{vehicles.}
    \colorhighlight{red}{0}{The}\colorhighlight{red}{0}{most}\colorhighlight{red}{0}{common}\colorhighlight{red}{0}{type}\colorhighlight{red}{0}{of}\colorhighlight{red}{0}{motorsport}\colorhighlight{red}{0}{is}\colorhighlight{red}{60.0}{car}\colorhighlight{red}{74.11764706}{racing}\colorhighlight{red}{0}{which}\colorhighlight{red}{0}{includes}\colorhighlight{red}{0}{events}\colorhighlight{red}{0}{such}\colorhighlight{red}{0}{as}\colorhighlight{red}{0}{Formula}\colorhighlight{red}{0}{One}\colorhighlight{red}{0}{NASCAR}\colorhighlight{red}{0}{and}\colorhighlight{red}{0}{rallying.}
    & 
    Motorsports is a sport that uses vehicles for competition.
    \colorhighlight{red}{0}{Many}\colorhighlight{red}{63.52941176}{people}\colorhighlight{red}{66.66666666}{disagree}\colorhighlight{red}{0}{because}\colorhighlight{red}{0}{they}\colorhighlight{red}{81.0}{say}\colorhighlight{red}{0}{that}\colorhighlight{red}{0}{it}\colorhighlight{red}{0}{is}\colorhighlight{red}{0}{dangerous}\colorhighlight{red}{0}{and}\colorhighlight{red}{0}{cruel}\colorhighlight{red}{0}{to}\colorhighlight{red}{0}{animals}\colorhighlight{red}{0}{but}\colorhighlight{red}{0}{what}\colorhighlight{red}{58.947368420000004}{do}\colorhighlight{red}{76.84210526000001}{you}\colorhighlight{red}{0}{think.}
    \colorhighlight{red}{70.52631577999999}{You}\colorhighlight{red}{0}{can}\colorhighlight{red}{60.95238096}{share}\colorhighlight{red}{0}{your}\colorhighlight{red}{0}{opinion}\colorhighlight{red}{0}{on}\colorhighlight{red}{0}{social}\colorhighlight{red}{0}{network}\colorhighlight{red}{0}{like}\colorhighlight{red}{0}{facebook}\colorhighlight{red}{0}{or}\colorhighlight{red}{70.52631577999999}{you}\colorhighlight{red}{0}{can}\colorhighlight{red}{0}{comment}\colorhighlight{red}{0}{this}\colorhighlight{red}{0}{forum.}
    &
    Motorsports is a sport that uses vehicles for competition.    \colorhighlight{green}{50}{\textbf{Races}}\colorhighlight{red}{0}{can}\colorhighlight{red}{0}{be}\colorhighlight{red}{0}{held}\colorhighlight{red}{0}{on}\colorhighlight{red}{0}{land}\colorhighlight{red}{64.21052632}{water}\colorhighlight{red}{69.0}{ice}\colorhighlight{red}{0}{or}\colorhighlight{red}{60.0}{snow}\colorhighlight{red}{0}{and}\colorhighlight{red}{0}{may}\colorhighlight{red}{0}{be}\colorhighlight{red}{0}{based}\colorhighlight{red}{0}{on}\colorhighlight{red}{0}{speed}\colorhighlight{red}{0}{distance}\colorhighlight{red}{0}{or}\colorhighlight{red}{0}{performance}\colorhighlight{red}{0}{criteria}\colorhighlight{red}{0}{such}\colorhighlight{red}{0}{as}\colorhighlight{red}{62.10526316}{endurance}\colorhighlight{red}{0}{or}\colorhighlight{red}{0}{durability.}
    \colorhighlight{red}{0}{Races}\colorhighlight{red}{0}{may}\colorhighlight{red}{0}{also}\colorhighlight{red}{0}{involve}\colorhighlight{red}{0}{other}\colorhighlight{red}{0}{forms}\colorhighlight{red}{0}{of}\colorhighlight{red}{77.0}{human}\colorhighlight{red}{0}{propelled}\colorhighlight{red}{0}{vehicle}\colorhighlight{red}{0}{or}\colorhighlight{red}{0}{animal.}
    \colorhighlight{green}{50}{\textbf{Racers}}\colorhighlight{red}{0}{usually}\colorhighlight{red}{57.142857140000004}{compete}\colorhighlight{red}{0}{against}\colorhighlight{red}{0}{each}\colorhighlight{red}{0}{other}\colorhighlight{red}{0}{but}\colorhighlight{red}{0}{races}\colorhighlight{red}{0}{may}\colorhighlight{red}{0}{also}\colorhighlight{red}{0}{be}\colorhighlight{red}{0}{against}\colorhighlight{red}{0}{time.}
    &  
    Motorsports is a sport that uses vehicles for competition.
    \colorhighlight{green}{50}{\textbf{Rallying}}\colorhighlight{red}{0}{and}\colorhighlight{red}{0}{rallycross}\colorhighlight{red}{0}{are}\colorhighlight{red}{0}{popular}\colorhighlight{red}{0}{forms}\colorhighlight{red}{0}{of}\colorhighlight{red}{0}{motorsport.}
    \colorhighlight{red}{73.68421052000001}{Car}\colorhighlight{red}{51.764705879999994}{racing}\colorhighlight{red}{0}{can}\colorhighlight{red}{78.94736842}{take}\colorhighlight{red}{0}{many}\colorhighlight{red}{0}{different}\colorhighlight{red}{0}{forms}\colorhighlight{red}{0}{such}\colorhighlight{red}{0}{as}\colorhighlight{red}{54.736842100000004}{drag}\colorhighlight{red}{51.764705879999994}{racing,}\colorhighlight{red}{0}{dirt}\colorhighlight{red}{0}{track}\colorhighlight{red}{51.764705879999994}{racing,}\colorhighlight{red}{0}{rally}\colorhighlight{red}{51.764705879999994}{racing,}\colorhighlight{red}{0}{stock}\colorhighlight{red}{73.68421052000001}{car}\colorhighlight{red}{51.764705879999994}{racing,}\colorhighlight{red}{0}{touring}\colorhighlight{red}{73.68421052000001}{car}\colorhighlight{red}{51.764705879999994}{racing,}\colorhighlight{red}{74.44444444}{sports}\colorhighlight{red}{73.68421052000001}{car}\colorhighlight{red}{51.764705879999994}{racing,}\colorhighlight{red}{63.0}{open}\colorhighlight{red}{52.631578940000004}{wheel,}\colorhighlight{red}{51.764705879999994}{racing}\colorhighlight{red}{0}{and}\colorhighlight{red}{0}{kart}\colorhighlight{red}{51.764705879999994}{racing.}
    \colorhighlight{green}{50}{\textbf{Understanding}}\colorhighlight{red}{0}{the}\colorhighlight{red}{0}{differences}\colorhighlight{red}{0}{between}\colorhighlight{red}{0}{these}\colorhighlight{red}{0}{forms}\colorhighlight{red}{0}{of}\colorhighlight{red}{74.11764706}{racing}\colorhighlight{red}{0}{as}\colorhighlight{red}{0}{well}\colorhighlight{red}{0}{as}\colorhighlight{red}{0}{the}\colorhighlight{red}{74.11764706}{racing}\colorhighlight{red}{57.49999999999999}{world}\colorhighlight{red}{0}{in}\colorhighlight{red}{0}{general}\colorhighlight{red}{0}{is}\colorhighlight{red}{0}{difficult}\colorhighlight{red}{0}{if}\colorhighlight{red}{65.2631579}{you}\colorhighlight{red}{74.7368421}{do}\colorhighlight{red}{0}{not}\colorhighlight{red}{70.0}{follow}\colorhighlight{red}{74.11764706}{racing}\colorhighlight{red}{0}{closely.}
    \\\bottomrule
    Confidence Score  & 0.6363 & 0.8692 & 0.9983 & 0.9995 \\
\end{tabular}
\caption{Example for a single prompt with 4 different configurations.
Red highlights indicate sensorimotor words (which may occur in any configuration, as a key can be generated even without watermark), and green with bold font highlights acrostics. 
The configurations with only acrostics and the combined version provide a confidence score above the significance threshold.}
\label{tab:examples_colored2}
\end{table*}

\subsection{Watermark Detection} \label{detection}
The detection of the watermark works like the inverse of the construction. 
First, the key used to encode the stylometric features needs to be recovered.
A block of the text is taken, in our case a sentence, and classified the same way with our semantic labels for the zero shot classifier to get the key values. 
The next block is then analysed based on this key for its stylometric features.

\label{def:inj}
The text we analyse for watermarks consists of $n$ sentences. 
For each sentence $s_i$, with $i>1$, we calculate $X_i$ by getting the average sensorimotor value depending on our key (i.e., sensorimotor category) of all words in $s_i$ given by the database \cite{lynott_lancaster_2020}.
We next calculate the z-score of each sentence, $z_i = \frac{X_i - \mu}{\sigma}$. To combine all $z_i$, we use Stouffer's method $Z = \frac{\sum{z_i}}{\sqrt{n}}$ to get a z-score for the whole text \cite{stouffer_american_1949,mullen_basic_1982}. This is then converted to a probability value $p_s$ using a standard normal distribution.

To calculate the probability of the acrostic feature  $p_a$, we make the assumption that all letters are equally distributed and independent of each other to appear at the beginning of a sentence. We count how many acrostica $a_i$ of all sentences are correct as $k$ and use the binomial probability mass function to calculate $p_a = \binom{n}{k}(\frac{1}{26})^{k} (1 - \frac{1}{26})^{n - k} $

The overall p-value then is $P = p_s * p_a$.

\section{Evaluation}\label{evaulation}

In this section we describe the hardware and implementation details, as well as the conducted experiments.
These include an ablation study to individually assess the contribution of the stylometric features.
Finally, we assess the resilience of our approach via a translation attack.

\subsection{Hardware and Implementation.} 

For the experimental setup, Mistral 7B~\cite{jiang_mistral_2023} was chosen as the model for our experiment.  We point out that our implementation is model independent and Mistral was only chosen because of its good performance and its open weight license. The experiments were conducted on an Nvidia RTX 4090 GPU and used GPTQ~\cite{frantar_gptq_2023} for 4-bit quantization to reduce video memory footprint. To manipulate the weights, the logits processor of Hugging Face~\cite{wolf_huggingfaces_2020} was used. For the zero classification for the key generation, we used Bart~\cite{lewis_bart_2019} trained on the MultiNLI dataset \cite{williams_broad-coverage_2018}. Refer to Section~\ref{construction} for more details. 

For detection, it is assumed that all features are independent of each other. Furthermore, acrostica were given the same probability across all letters. Refer to Section~\ref{detection} for more details. 

\subsection{Prompt Generation and Scoring}

A total of 156 sample prompts were generated with help of ChatGPT~\cite{openai_chat_2024a}. The ChatML format \cite{microsoft_how_2023} was used for prompting, and every prompt was given with the same hyperparameters with a fixed sampling seed for reproducibility. 
Replies were limited to a length of 25 sentences. 
We use a level of $\alpha = 0.05$ to then determine the statistical significance that a watermark can be recovered.

\subsection{Results}

In Table~\ref{tab:examples_colored2} an example illustrates the output of an unaltered response, together with the individual features and the final output of our approach on the right side (more examples can be found in the Appendix~\ref{tab:demo-examples}).
The first sentence is highlighted, as well as the words controlled via the acrostic feature and the sensorimotor words.
In the bottom row the confidence score is reported, with both the acrostica and the combined features are considered statistically significant.

\begin{figure*}[ht!]
    \vskip 0.2in
    \begin{center}
    \centerline{\includegraphics[width=0.8\textwidth]{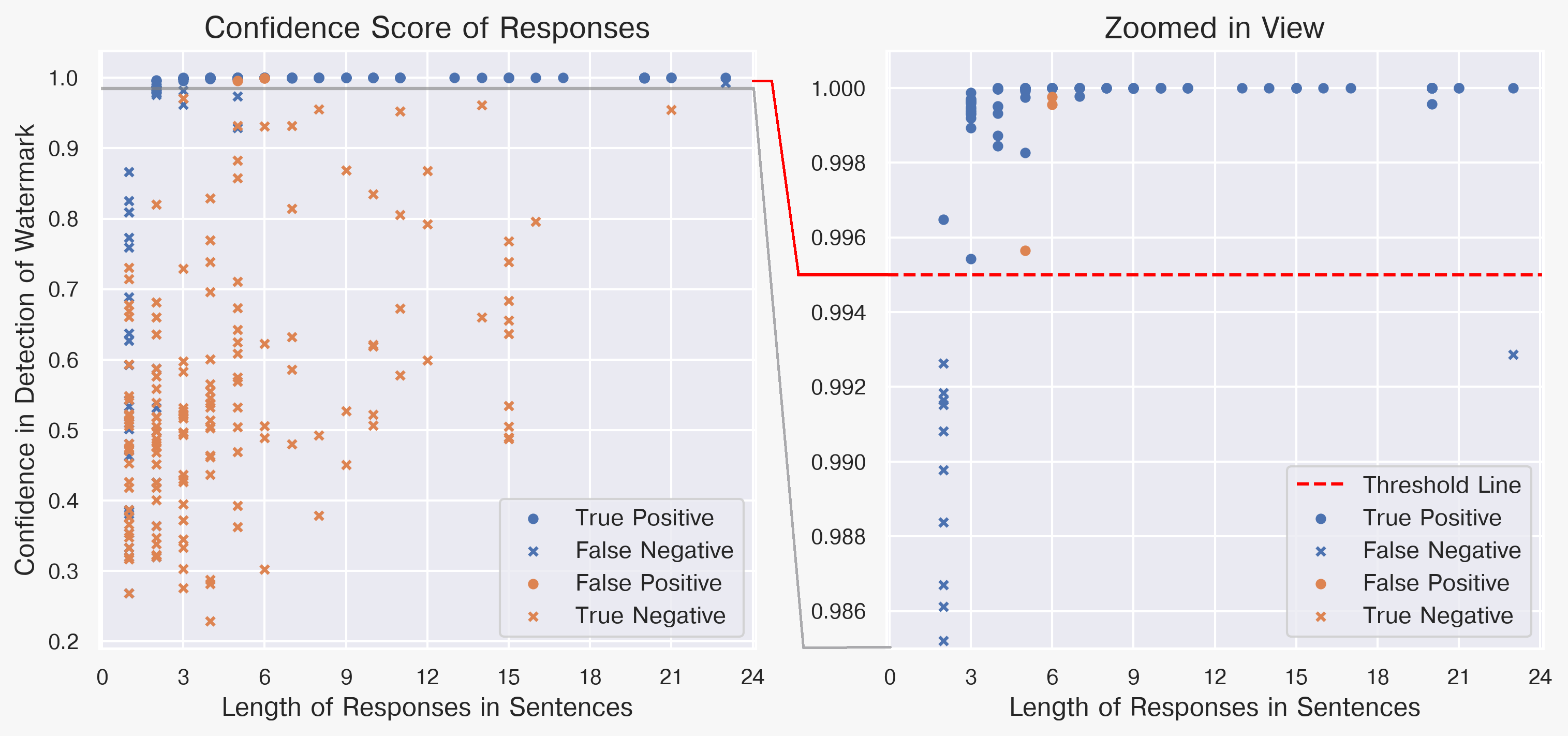}}
    \caption{Plot of the 156 prompts with unaltered (orange) and watermarked responses (blue), with a zoomed in chart on the right side to highlight the level of significance, with the length of the response in sentences as the x-axis. Dots above the threshold line are considered to be statistically significant. 
    There are no false positives for responses of more than 6 sentences.
    }
    \label{fig:plot_paper}
    \end{center}
    \vskip -0.2in
\end{figure*}

Figure \ref{fig:plot_paper} presents a scatter-plot of the results of all prompts with the length of the response, measured in number of sentences and the confidence score as axis.
Here, the unaltered version is compared with the combined stylometric watermark.
Dots above the threshold line represent statistically significant results for detecting a watermark.
Most results of the unaltered base results are centred at score between $0.1$ and $0.7$, resembling the form of a typical Gaussian distribution. 
However, six of them are in the upper part with a score higher than $0.9$. 
Out of those, three would have been classified as a false positive with our confidence limit $\alpha$ with a maximum sentence length of $6$. 
This is caused in all cases by the wrong detection of an acrostic that causes a high confidence accidentally. However, if this was still a too high false positive rate, the minimal text length could simply be increased.

The watermarked answers are clustered in two different spots depending on the numbers of sentences. 
The very short replies with under two sentences were clustered between $0.1$ and $0.9$. 
This is due to two sentences being the bare minimum for our watermark to start working, since one sentence is required to calculate the key and another sentence to compare and evaluate the watermark. 
The second group with more than three sentences in length are nearly all detected correctly, with only three false negatives. As expected, longer responses also have a minimally higher confidence score, but detection already works well with short responses.

\subsection{Ablation Study}

To further analyse the contribution  of the two considered features on the score, we apply the analysis individually.
Table~\ref{ablationstudy} contains the result of our watermark classification experiment for each feature alone and all combined. Only responses with more than 3 sentences were included, which leads us to having 164 replies. A visual representation can be found in Appendix~\ref{fig:ablation}.
They display a vastly different behaviour, and each of them being worse than the combined configuration. This difference grows even larger when an attack has happened, as can be seen through our cyclic translation experiment.

\begin{table}[t]
\centering
\caption{Results for the feature combination. Only responses more than 3 sentences long were included, yielding 164 replies. (TP = true positive rate, TN = true negative rate, FP = false positive rate, FN = false negative rate)}
\label{ablationstudy}
\begin{tabular}{@{}lcccc@{}}
\toprule
                 & TP      & TN      & FP     & FN \\ \midrule
Sensorimotor   &  0.07  &  0.48  &   0.0   & 0.45           \\
Acrostic         &  0.49  &  0.46  &   0.01  & 0.04      \\
\midrule
Both             &  0.51  &  0.46  &   0.02  & 0.02         \\
\bottomrule
\end{tabular}
\end{table}

\subsection{Resilience Against Attacks} \label{robustness}
To further evaluate the efficacy of stylometric watermarking, it is important to consider the nature and impact of different attack methods users could employ to tamper with the watermark. This subsection delves into possible attacks and evaluates our method on the most difficult one that users can employ with little effort.

\begin{figure*}[t]
    \vskip 0.2in
    \begin{center}
    \centerline{\includegraphics[width=0.8\textwidth]{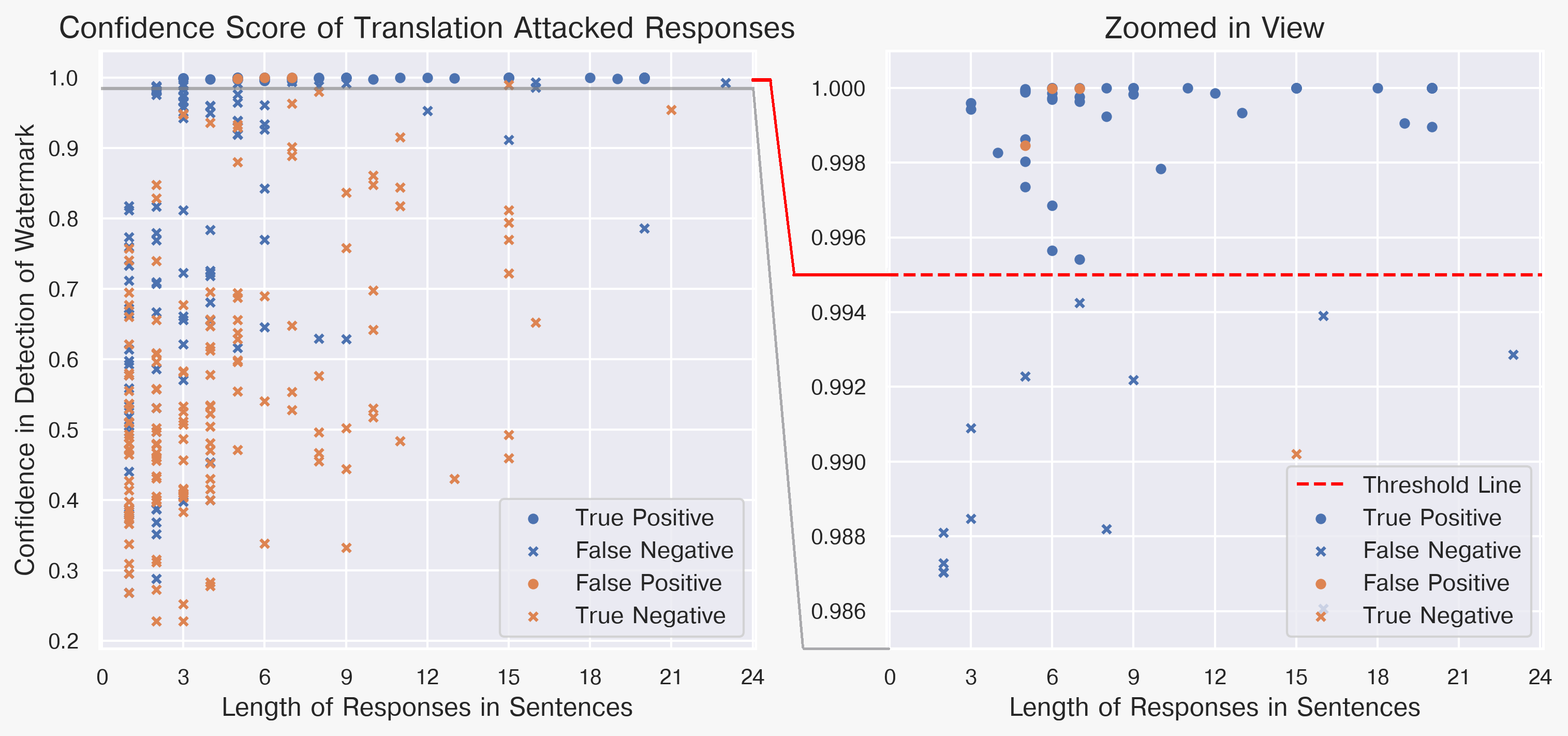}}
    \caption{Plot of the 156 prompts for normal and watermarked responses after being attacked by a cyclic translation. Even for this type of attack, the watermark was recoverable for the majority of prompts. For responses longer than 7 sentences, the attack was successful only in 9 cases.}
    \label{fig:attacked}
    \end{center}
    \vskip -0.2in
\end{figure*}

\paragraph{Types of Attacks and Their Implications.}
Various attacks such as cyclic translation, paraphrasing, and sentence shifting can have detrimental effects on the effectiveness of a watermark. For instance, the integrity could be compromised in several ways:
\begin{itemize}
    \item \textbf{Damaged Keys:} The attacks might lead to the distortion of keys embedded within the content, leading to wrong keys being used in detection and thus hinder the ability to detect the watermark.
    \item \textbf{Shifted Key Sentences:} The sections of the text generating the keys could be shifted in position, leading to the problems with finding the correct key for each watermarked section.
    \item \textbf{Missing Key Sentences:} A related version of the previous issue, with sentences containing the key might be omitted completely, rendering at least the following watermark useless since it cannot be detected any more.
\end{itemize}

However, if a resilient watermarking technique is employed using multiple features simultaneously, it becomes significantly harder to remove the watermark.

\paragraph{Experiment on Cyclic Translation Attacks.}
An experiment was conducted to assess the vulnerability of our watermarking features and method to translation attacks. This involved using Google Translate~\cite{google_cloud_2024} to convert English text into Spanish and then back into English. The process was repeated for all 312 text samples generated in the previous experiment. Refer to Appendix~\ref{tab:demo-translation} for an example.

As can be seen in Figure \ref{fig:attacked}, both our key and our stylometric features held up very well. The attack was unable to change the semantic classification of sentences to destroy enough keys.
The acrostic and sensorimotor features were also not removed by the attack to the extent to break the detection. The bound for detection moved from three sentences before to seven sentences for detection with similar confidence. Only nine responses managed to go undetected due to the attack.

\begin{table}[t]
\centering
\caption{Results of the experiment with the cyclic translation  attack, limited to responses with more than three sentences (151 replies). As expected, the false negative rate did increase, contributed mostly by short replies.}
\label{translation}
\begin{tabular}{@{}lcccc@{}}
\toprule
                 & TP      & TN      & FP     & FN \\ \midrule
Sensorimotor    &  0.04  & 0.48  & 0.0&  0.48\\
Acrostic          &  0.22 & 0.47  & 0.01& 0.30\\
\midrule
Both              &  0.28   & 0.46  & 0.02& 0.24 \\
\bottomrule
\end{tabular}
\end{table}

\section{Conclusion}
We presented a novel watermark method for large language models, in particular generative transformer models.
Out of the possible approaches for integrating watermarks in LLMs we opted for manipulating the probability directly when generating the tokens. 
One key aspect of the approach is to change these probabilities per sentence based on a key, which is dynamically derived directly from the generated text.
Out of many possible stylometric features, and we focused on two features for our experiments, the acrostic and sensorimotor words. 

We evaluate a range of prompts with and without applying our watermark and show that the method can differentiate responses with a very high degree of confidence with three or more sentences.
This result is satisfying given that the minimum theoretical number of sentences required by the approach is two.
The evaluation also includes possible attacks on a textual watermark, and the method is tested by attempting to damage or remove the watermark with a cyclic translation attack. 
We show that by increasing the required sentences to seven, our watermark is resilient to the attack and can still be robustly detected with high confidence.
Overall, we developed a resilient watermark for text that works with short text length, and does not require any expensive additional model training or an LLM for testing.
Since our method allows for many types of key generation and stylometric features, an exploration of their combinations will be part of the future work.

\bibliographystyle{unsrtnat}
\bibliography{zotero_references,references}  

\newpage
\appendix
\onecolumn
\section{Appendix}

\begin{table}[ht!]
\caption{Additional prompts showcased as an example. Each prompt has its base reply without any watermark, then the watermarked reply using only one feature at the time, and finally the reply using both features. Using both features at the same time usually leads to the highest strength watermark, but here the second prompt shows different behaviour. This is because we are not using large boosts for the model's weights, and this leads to the watermark with both features using only one acrostic, while the acrostica only watermark uses two. The longer a reply is allowed to get, the rare this case becomes.}
\tiny
\begin{tabular}{p{2.1cm}|p{3.0cm}|p{3.0cm}|p{3.0cm}|p{3.0cm}}
\toprule
prompt & base &  sensorimotor &  acrostic & both  \\
\midrule
 What is a watermark? & A watermark is any mark that appears on a digital image to indicate ownership.  Watermarks can be text, graphics, or both.The purpose of a watermark is to discourage unauthorized reproduction and/or distribution of an image. &  A watermark is any mark that appears on a digital image to indicate ownership. Watermarks can be text or graphics added into different parts of digital images, such as the corner or across one side of the image.Watermarking is often done with text, like name or logo, but graphics can be used just as effectively.The purpose of watermark is to prevent against copyright infringement or illegal use of digital image content. & A watermark is any mark that appears on a digital image to indicate ownership.  By default, you can enable this option in your site settings. But you can also disable it if you wish, for example, if you want to create a product with your own watermark and add it later manually. By default, you can enable this option in your site settings. & A watermark is any mark that appears on a digital image to indicate ownership. By law you can do this with text or art. Visually you can do it with text, art, shape, position, transparency and size. Many watermarks use text like "copyright 2014" or a logo. \\\midrule
 Confidence Score & 0.4555 & 0.5664 & 0.99996 & 0.99998 \\\midrule
 How do I write a good paper?  & A good paper should be well structured and include relevant information. It should also be easy to read and understand, with clear organization and logical flow of ideas.The introduction should clearly state the purpose of the paper, including the research question or problem being addressed. The methods section should describe in detail how the research was conducted, including any materials and procedures used. & A good paper should be well structured and include relevant information. It is important to clearly explain your methodology, results, discussion, and conclusion in order to communicate effectively with others in your field.It is also important to properly reference all sources used in your research, including data sets, software code, and literature.A good paper is clear, concise, and well organized. & A good paper should be well structured and include relevant information. Consider using headings and subheadings to organize your thoughts, and make sure to provide enough detail so that your reader can understand your argument clearly. You may also want to consider incorporating evidence from other sources to support your claims. & A good paper should be well structured and include relevant information. Consider what you want to achieve with it, what target group you are writing it for and what you want them to know after reading it. Next you can structure your paper into an introduction, body text and conclusion, while also considering which format you use. \\\midrule
 Confidence Score & 0.6224 &  0.7812 & 0.9947 & 0.9870 \\
 \bottomrule
\end{tabular}

\label{tab:demo-examples}
\end{table}

\begin{table}[ht!]
\caption{Example of an acrostic written by Arnold Schwarzenegger in a letter to the members of the California State Assembly, which has been used in literature for illustration before \citep{stein2014generating}.}
\centering
\begin{tabular}{@{}p{\textwidth}@{}}
\toprule
To the Members of the California State Assembly: \\
\\
I am returning Assembly Bill 1176 without my signature. \\
\\
\textbf{F}\,or some time now I have lamented the fact that major issues are overlooked while many \\
\textbf{u}\,nnecessary bills come to me for consideration. Water reform, prison reform, and health \\
\textbf{c}\,are are major issues my Administration has brought to the table, but the Legislature just \\
\textbf{k}\,icks the can down the alley. \\
\\
\textbf{Y}\,et another legislative year has come and gone without the major reforms Californians \\
\textbf{o}\,verwhelmingly deserve. In light of this, and after careful consideration, I believe it is\\
\textbf{u}\,nnecessary to sign this measure at this time. \\
\\
Sincerely, \\
Arnold Schwarzenegger \\
\bottomrule
\end{tabular}
\label{tab:acrostic-schwarz}
\end{table}

\begin{figure}[ht!]
    \vskip 0.2in
    \begin{center}
    \centerline{\includegraphics[width=\textwidth]{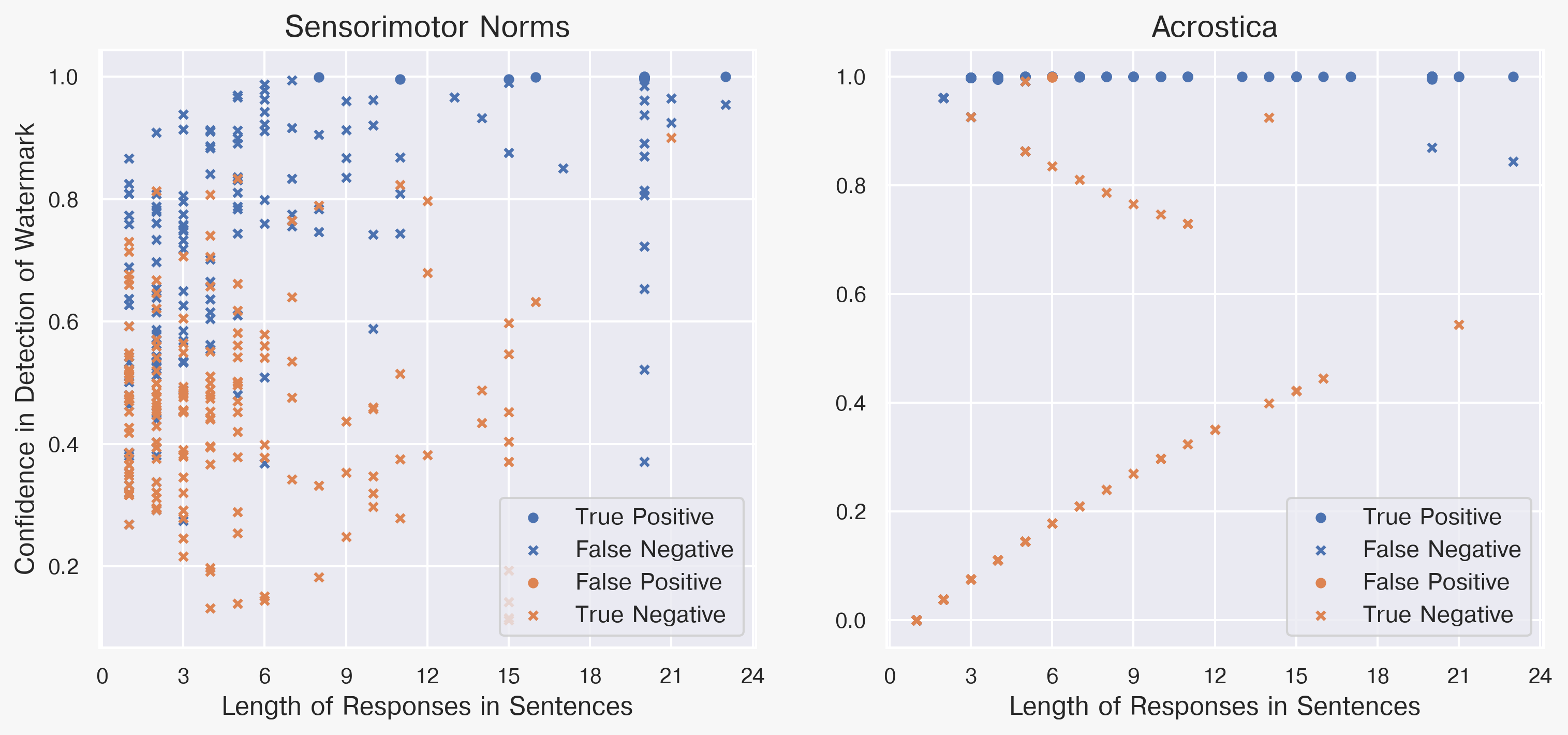}}
    \caption{Comparison of the individual contribution of the two feature types. On the left the results for the sensorimotor features are shown and the results of the acrostica on the right,}
    \label{fig:ablation}
    \end{center}
    \vskip -0.2in
\end{figure}

\begin{table}[ht!]
\caption{An example from our dataset of how a cyclic translation attack looks like, translating the original outputs into Spanish and back into English using Google Translate~\cite{google_cloud_2024}. For output with many sentences like here, the watermark stays intact enough to keep the high confidence score.}
\tiny
\begin{tabular}{p{1.7cm}|p{7.0cm}|p{7.0cm}}
\toprule
 \multicolumn{3}{c}{Explain the concept of artificial intelligence and its potential impacts} \\
\midrule
 & Base &  Watermarked  \\
\midrule
  Original Output& Artificial Intelligence (AI) is a term used to describe various technologies that enable computers to perform tasks that normally require human intelligence, such as understanding language, recognizing objects in images, making decisions based on uncertain data, or learning from experience. AI has the potential to revolutionize many industries, including healthcare, finance, transportation, and manufacturing, by enabling machines to automate complex tasks and make more informed decisions than humans can. However, there are also concerns about the ethical implications of AI, particularly regarding issues like bias, privacy, and security. Overall, the impact of AI will likely be both positive and negative, depending on how it is implemented and regulated. & Artificial Intelligence (AI) is a term used to describe various technologies that enable computers to perform tasks that normally require human intelligence, such as understanding language, recognizing objects in images, making decisions based on uncertain data, or learning from experience. GPT-3 is an AI model trained on large amounts of text data to generate human-like responses to prompts. GPT-3 is capable of generating human-sounding content on a variety of topics, but it cannot understand context or make logical deductions on its own. GPT-3 is not capable of real critical thinking, problem solving, reasoning, creativity, emotional awareness, moral judgment, self awareness, social skills or decision making on its own. Zero-shot prompts like this one do not exist in nature. Research into the effects of AI is ongoing, but some experts have raised concerns about the potential impact of AI on jobs, privacy, ethics, and national security. GPT-3 is still far off human level intelligence, but research continues to improve the performance of machine learning models over time. \\ \midrule
  Confidence Score &  0.53238325011  &  0.99999999984  \\ \midrule
 Spanish Output & Inteligencia artificial (IA) es un término utilizado para describir diversas tecnologías que permiten a las computadoras realizar tareas que normalmente requieren inteligencia humana, como comprender el lenguaje, reconocer objetos en imágenes, tomar decisiones basadas en datos inciertos o aprender de la experiencia. La IA tiene el potencial de revolucionar muchas industrias, incluidas la atención médica, las finanzas, el transporte y la manufactura, al permitir que las máquinas automaticen tareas complejas y tomen decisiones más informadas que los humanos. Sin embargo, también existen preocupaciones sobre las implicaciones éticas de la IA, particularmente en cuestiones como el sesgo, la privacidad y la seguridad. En general, el impacto de la IA probablemente será tanto positivo como negativo, dependiendo de cómo se implemente y regule. & Inteligencia artificial (IA) es un término utilizado para describir diversas tecnologías que permiten a las computadoras realizar tareas que normalmente requieren inteligencia humana, como comprender el lenguaje, reconocer objetos en imágenes, tomar decisiones basadas en datos inciertos o aprender de la experiencia. GPT-3 es un modelo de IA entrenado con grandes cantidades de datos de texto para generar respuestas similares a las humanas a las indicaciones. GPT-3 es capaz de generar contenido que suena humano sobre una variedad de temas, pero no puede comprender el contexto ni hacer deducciones lógicas por sí solo. GPT-3 no es capaz de tener pensamiento crítico real, resolución de problemas, razonamiento, creatividad, conciencia emocional, juicio moral, autoconciencia, habilidades sociales o toma de decisiones por sí solo. Los avisos de disparo cero como este no existen en la naturaleza. La investigación sobre los efectos de la IA está en curso, pero algunos expertos han expresado su preocupación sobre el impacto potencial de la IA en el empleo, la privacidad, la ética y la seguridad nacional. GPT-3 todavía está lejos del nivel de inteligencia humana, pero la investigación continúa para mejorar el rendimiento de los modelos de aprendizaje automático con el tiempo. \\\midrule
 Cyclic Translation Output & Artificial intelligence (AI) is a term used to describe several technologies that allow computers to perform tasks that normally require human intelligence, such as understanding language, recognizing objects in images, making decisions based on uncertain data or learning from experience.The AI has the potential to revolutionize many industries, including medical care, finance, transport and manufacturing, allowing machines to automate complex tasks and make more informed decisions than humans. However, there are also concerns about implicationsEthics of AI, particularly with respect to issues such as bias, privacy and security. In general, the impact of AI will probably be positive and negative, depending on how it is implemented and regulated. & Artificial intelligence (AI) is a term used to describe several technologies that allow computers to perform tasks that normally require human intelligence, such as understanding language, recognizing objects in images, making decisions based on uncertain data or learning from experience.GPT-3 is a model of the trained in large amounts of text data to generate similar responses to humans to the indications.GPT-3 is capable of generating content that sounds human in a variety of topics, but cannot understand the contextor make logical deductions by themselves.of zero shooting as this does not exist in nature. The research on the effects of AI is in progress, but some experts have raised concerns about the potential impact of AI on jobs, privacy, ethics and national security.GPT-3 is still far from intelligence at a human level, but research continues to improve the performance of automatic learning models over time. \\ \midrule
 Confidence Score & 0.57793936438 & 0.99999968547 \\
 \bottomrule
\end{tabular}
\label{tab:demo-translation}
\end{table}

\begin{table}[ht!]
\caption{Our labels we created with the help of ChatGPT~\cite{openai_chat_2024a}. Our generalized semantic zero shot classification can use them to construct our key because they cover a broad range of semantic meanings.}
\centering
\begin{tabular}{>{\raggedright}p{3.5cm}p{6cm}p{3.5cm}}
\toprule
\textbf{Feature Type} & \textbf{Semantic Label}& \textbf{Corresponding Token} \\
\midrule
Acrostica & Scientific Concepts & A \\
 & Technical Explanations & B \\
 & Historical Context & C \\ 
 & Cultural Insights & D \\
 & Environmental Issues & E \\
 & Health and Medicine & F \\
 & Technological Developments & G \\
 & Economic Theories & H \\
 & Political Analysis & I \\
 & Philosophical Concepts & J \\
 & Educational Methods & K \\
 & Psychological Theories & L \\
 & Artistic Movements & M \\
 & Literary Analysis & N \\
 & Global Events & O \\
 & Culinary Traditions & P \\
 & Mathematical Concepts & Q \\
 & Physical Principles & R \\
 & Astronomical Discoveries & S \\
 & Geographical Information & T \\
 & Social Dynamics & U \\
 & Legal Interpretations & V \\
 & Business Strategies & W \\
 & Sports and Fitness & X \\
 & Linguistic Features & Y \\
 & Techniques in Science and Technology & Z \\
\addlinespace \midrule
Sensorimotor Words & Science and Technology & Auditory \\
& Health and Environmental Issues & Gustatory \\
& Arts, Culture, and History & Haptic \\
& Economic and Political Analysis & Interoceptive \\
& Philosophy and Psychology & Olfactory \\
& Education and Learning Methods & Visual \\
& Global and Social Dynamics & Foot/Leg \\
& Legal and Ethical Discussions & Hand/Arm \\
& Business and Management & Head \\
& Sports, Fitness, and Recreation & Mouth \\
& Language and Literature & Torso \\
\addlinespace
\bottomrule
\end{tabular}
\label{table:labels}
\end{table}

\clearpage
\begin{longtable}[ht!]{p{0.5\linewidth}}
\caption{All 156 prompts that were used in the experiments.}\\
\toprule
\textbf{Prompt} \\
\midrule
What is the capital of France? \\
Explain the process of photosynthesis.  \\
What are the basic rules of chess?  \\
Describe the water cycle.  \\
Summarize the plot of 'Romeo and Juliet'. \\
What is Newton's first law of motion?  \\
How do you make a simple omelette?9 \\
List the seven continents.\\
What are the primary colors?\\
Explain the concept of supply and demand in economics. \\
What is the function of the heart in the human body?  \\
How does a compass work?  \\
Define the term 'ecosystem'. \\
What is the boiling point of water? \\
Describe how a computer works. \\
What are the planets in our solar system? \\
Explain the basics of HTML.  \\
Who was Albert Einstein?  \\
What is the Pythagorean theorem?  \\
How do you calculate the area of a circle?  \\
What is photosynthesis?  \\
Describe the life cycle of a butterfly. \\
What is the Declaration of Independence?  \\
How do you change a flat tire?  \\
What is the process of evaporation?  \\
Explain the rules of soccer.  \\
What is the significance of the Great Wall of China? \\
How do solar panels generate electricity?  \\
What causes earthquakes?  \\
Describe the structure of DNA. \\
What is a black hole?  \\
How do airplanes fly?  \\
Explain the concept of gravity.  \\
What is global warming?  \\
Describe the human digestive system.\\
What is the Renaissance? \\
How does the stock market work?  \\
What are renewable energy sources?  \\
Explain the basic principles of democracy. \\
What is virtual reality?  \\
How do vaccines work?  \\
What is the internet? 5 \\
Describe the process of making chocolate.  \\
What is a haiku?  \\
How do bees make honey? \\
What is the theory of relativity?\\
Explain the function of the lungs. \\
What is a balanced diet?  \\
Describe the phases of the moon.  \\
What is artificial intelligence?  \\
How does a refrigerator work?  \\
Explain the concept of time zones.  \\
What is the significance of the Mona Lisa? \\
How is glass made? \\
What are the causes of World War II?  \\
Explain the importance of biodiversity. \\
What is the structure of an atom?  \\
How does a camera work?  \\
What is mindfulness?  \\
Describe the process of recycling. \\
What causes the seasons to change?\\
Describe the French Revolution.\\
What is quantum mechanics?\\
How do you solve a quadratic equation?\\
What is the significance of the Berlin Wall?\\
Explain the process of fermentation.\\
What is the human genome project?\\
How does the internet work?\\
What are the symptoms of diabetes?\\
Describe the process of cell division.\\
What is the theory of evolution?\\
How do you make bread?\\
What are the major religions of the world?\\
Explain the concept of inflation in economics.\\
What is the significance of the Magna Carta?\\
How does a microwave oven work?\\
What is the structure of the United Nations?\\
Describe the American Civil War.\\
What causes tsunamis?\\
How do you write a business plan?\\
What is the importance of the Rosetta Stone?\\
Explain the principles of non-violent resistance.\\
What is a supernova?\\
How do you calculate compound interest?\\
Describe the process of photosynthesis in detail.\\
What is the impact of climate change on oceans?\\
Explain the concept of blockchain.\\
What is the importance of the Silk Road?\\
How do antibiotics work?\\
Describe the Greek pantheon of gods.\\
What is dark matter?\\
How do you make a website?\\
Explain the process of nuclear fusion.\\
What is the history of the internet?\\
Describe the principles of classical architecture.\\
What causes volcanic eruptions?\\
How do you create a budget?\\
What is the importance of the Suez Canal?\\
Explain the concept of artificial selection.\\
What is the difference between mitosis and meiosis?\\
Describe the Crusades.\\
What is the formula for calculating velocity?\\
How does a sewing machine work?\\
What is the significance of the Dead Sea Scrolls?\\
Explain the basics of cryptocurrency.\\
What is the function of the kidneys in the human body?\\
Describe the battle of Gettysburg.\\
What causes global economic crises?\\
How do you make a pizza?\\
What is the theory behind black holes?\\
Describe the structure and function of the brain.\\
What is the significance of Machu Picchu?\\
How does a light bulb work?\\
Explain the concept of the Big Bang Theory.\\
What is the role of the World Health Organization?\\
Describe the process of making wine.\\
What is the importance of quantum computing?\\
How do you improve your memory?\\
What is the story behind the Trojan War?\\
Describe the function of the immune system.\\
What causes lightning and thunder?\\
How do you calculate BMI (Body Mass Index)?\\
What is the significance of Stonehenge?\\
Explain the concept of virtual private networks (VPNs).\\
What is the difference between renewable and non-renewable energy?\\
Describe the process of making beer.\\
What is the significance of the Panama Canal?\\
How does a GPS system work?\\
What is the history of the Olympic Games?\\
Explain the concept of machine learning.\\
What is the importance of the Gutenberg Bible?\\
How do you make a campfire?\\
What causes addiction?\\
Describe the process of making cheese.\\
What is the significance of the Hagia Sophia?\\
How does Bluetooth technology work?\\
What is the history of cryptography?\\
Explain the concept of sustainable development.\\
What is the importance of the International Space Station?\\
How do you make chocolate chip cookies?\\
What causes allergies?\\
Describe the process of making soap.\\
What is the significance of the Colosseum in Rome?\\
How does nuclear power work?\\
What is the history of the English language?\\
Explain the concept of the electoral college in the United States.\\
What is the importance of the Human Rights Declaration?\\
How do you make a smoothie?\\
What causes migraines?\\
Describe the process of making tea.\\
What is the significance of the Taj Mahal?\\
How does solar energy work?\\
What is the history of the Roman Empire?\\
Explain the concept of photosynthesis in algae.\\
What is the importance of the Great Barrier Reef?\\
How do you make a latte?\\
What causes insomnia?\\
Describe the process of making sushi.\\
What is the significance of the Great Pyramid of Giza?\\
How does wind energy work?\\
What is the history of the Vikings?\\
Explain the concept of genetic engineering.\\
What is the importance of the Mona Lisa in the art world?\\
\bottomrule
\end{longtable}

\end{document}